# Lip Localization and Viseme Classification for Visual Speech Recognition

**Salah Werda, Walid Mahdi and Abdelmajid Ben Hamadou**
MIRACL: Multimedia Information systems and Advanced Computing Laboratory
Higher Institute of Computer Science and Multimedia, Sfax, Tunisia
salah.werda@yahoo.fr; [walid.mahdi, benhamadou.abdelmajid] @isims.rnu.tn

**Abstract:**
*The need for an automatic lip-reading system is ever increasing. Infact, today, extraction and reliable analysis of facial movements make up an important part in many multimedia systems such as videoconference, low communication systems, lip-reading systems. In addition, visual information is imperative among people with special needs. We can imagine, for example, a dependent person ordering a machine with an easy lip movement or by a simple syllable pronunciation. Moreover, people with hearing problems compensate for their special needs by lip-reading as well as listening to the person with whome they are talking.*
*We present in this paper a new approach to automatically localize lip feature points in a speaker's face and to carry out a spatial-temporal tracking of these points. The extracted visual information is then classified in order to recognize the uttered viseme (visual phoneme). We have developed our Automatic Lip Feature Extraction prototype (ALiFE). Experiments revealed that our system recognizes 72.73% of French Vowels uttered by multiple speakers (female and male) under natural conditions.*

**Keywords:** *visual information, lip-reading system, Human-Machine interaction, lip, spatial-temporal tracking.*

## 1. Introduction

Visual speech cues play an important role in Human speech perception, especially in noisy environments. Phenomena such as the perceptive illusions of McGurk [4] (speakers confronted to an auditory stimuli /ba/ and a visual stimuli /ga/ perceive the stimuli /da/) or the "Cocktail party" effect (attention centered on a special speaker surrounded by multiple speakers discoursing at the same time) show the significance of visual information in speech perception.

In this context, many works in the literature, from the oldest [1] until the most recent ones [2], [3] and [23] have proved that movements of the mouth can be used as one of the speech recognition channels. Recognizing the content of speech based on observing the speaker's lip movements is called 'lip-reading'. It requires converting the mouth movements to a reliable mathematical index for possible visual recognition.

It is around this thematic that our ALiFE (Automatic Lip Feature Extraction) system appears. ALiFE allows visual speech recognition from a video locution sequence. More precisely, it implements our approach which is composed of three steps: First, it proceeds by localizing the lips and tracking them. Secondly, it extracts precise and pertinent visual features from the speaker's face. Finally, the extracted features are used for visemes (visual phoneme) classification and recognition. The ALiFE beta version system, presented in this paper, covers the totality of the visual speech recognition steps shown in figure1.

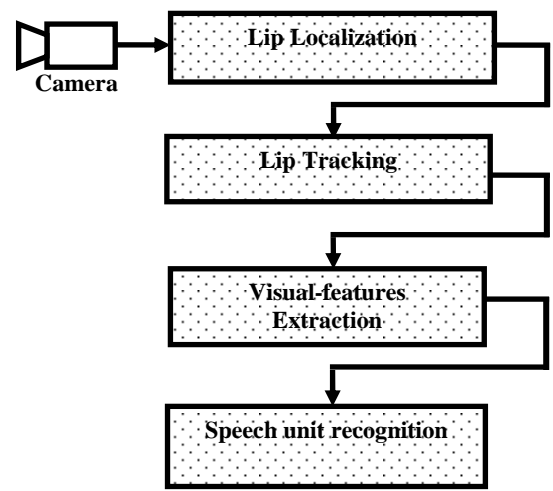

**Figure 1.** Overview of the complete ALiFE System for speech recognition

In section (2) we present an overview on visual speech recognition and labial segmentation methods currently proposed in the literature. Section (3) details out our lip localization and lip tracking methods. In section (4), we present the different features which will be used for the recognition stage and the different stage of our ALiFE viseme classification and recognition system. In section (5), we evaluate: First, our lip tracking method by comparing it with other tracking approaches. Secondly our ALiFE system for the visual recognition of each viseme present in our corpus. Rates of viseme recognition as well as a matrix of confusion between these visemes will be shown. We conclude the paper with a summary of the current state of our work and future.

## 2. Visual Speech Recognition and Labial Segmentation Methods: an overview

The human perception of the world is inherently multi-sensory since the information provided is multimodal. The perception of a spoken language is not an exception. Beside the auditory information, there is visual speech information as well, provided by the facial movements as a result of moving the articulators during speech production [24]. The complementary nature of visual and auditory percepts in the speech comprehension pushed several research teams on Automatic Speech Recognition (ASR) to focalize on visual channel. In the following sub-sections, we discuss the different methods evoked in the literature about labial segmentation and Visual Speech Recognition systems.

### 2.1 Labial Segmentation Methods

Several research works stressed their objectives in the research on automatic and semiautomatic methods for the extraction of visual indices, necessary to recognize visual speech (lip-reading) [1], [6], [10] and [23].

Two types of approaches have been used for lip-reading depending on the descriptors they use for the recognition of the visemes:

- The low-level approach (Image-based approaches) [5] and [6], controlled by data, use directly the image of the mouth region. This approach supposes that the lip pixels have a different feature compared to the ones of skin pixels. Theoretically, the segmentation can therefore be done while identifying and separating the lips and skin classes. In practice, methods of this type allow rapid locations of the interest zones and make some very simple measures of it (width and height of the lips, for example). However, they do not permit to carry out a precise detection of the lip edges.

- The high level approach (Model-based approaches) [7], [8], [9] and [2], which is directed by physical extraction distance, uses a model. For example, we can mention the active edges, which were widely used in lip segmentation. These approaches also exploit the pixel information of the image, but they integrate regularity constraints. The big deformability of these techniques allows them to be easily adapted to a variety of forms. This property is very interesting when it is a matter of segmenting objects whose form cannot be predicted in advance (sanguine vessels, clouds...), but it appears more as a handicap when the object structure is already known (mouth, face, hand...).

### 2.2 Visual Speech Recognition System

Most approaches for speech modeling using features are based on modeling techniques that are widely used in acoustic speech recognition.

Template matching of static images is a very simple method for visual speech recognition. It is the comparison of features extracted from static images with stored templates [24]. Such approaches ignore the temporal evolution of features and assume that an image, representing the important features of a given utterance, can be captured from the image sequence.

Petajan [1] compared the extracted sequence of geometric parameters with stored templates of feature sequences. No time wrapping was used to align the sequences. His system [25] was later extended by vector quantization and dynamic time warping (DTW). The quantized images were used in a DTW algorithm to find the best path between the unknown sequence and the stored template. However some of the disadvantages of DTW based speech modeling techniques are the distance measures between the features. Such measures don't consider their feature distribution and the temporal modeling which is not very specific and often is based on heuristic penalties. Petajan's system was extended by Goldschen [26] to a continuous speechreading system using discrete HMM (Hidden Markov Model). Distinct viseme groups were then determined using a HMM similarity metric [14] and a clustering algorithm.

The speechreading system reported by Bregler et al. [27] used a modular time-delay neural network (TDNN) which consists of an input layer, a hidden layer, and a phone state layer. The network was trained by back-propagation. Bergler et al. [27] have described another connectionist approach for combining acoustic and visual information into hybrid Multi layer Perceptron MLP/HMM speech recognition system. Given the audio-visual data MLP is trained to estimate the posterior probabilities of the phonemes. The likelihoods are obtained from the posterior probabilities and used as the emission probabilities for the HMM.

Finally, Nankaku and Tokuda, use the continuous density HMM method [28]. The ordinary approach in the conventional normalization is to provide a criterion independently of the HMM, and to apply normalization before learning. In their approach, normalization by the ML (Maximum Likelihood) criterion is considered. Normalized training is proposed such a way that the normalization processes for elements such as the position, size, inclination, mean brightness, and contrast of the lips are integrated with the training of the model.

## 2.3 Discussion

Potamianos [10] and Matthews [11] proved that image approaches allowed better performances in terms of recognition rates than image-based approaches in different conditions. On the other hand for Eveno [12] [13], the most promising methods in labial segmentation, are those model-based approaches because they are based on lip models. He proceeds by detecting the full contour of lips, but it is necessary to doubt on the interest of the extraction of the labial contours in their entirety for the recognition stage. According to speech specialists [2], the pertinent features of verbal communication expression are: the heights, widths and inter-labial surface. From this interpretation we notice that it will be judicious to opt for an extraction method of these features based on the detection and the tracking of some "Points Of Interest" (POI) sufficient to characterize labial movements. Therefore, the problem of labial segmentation is to detect some POI on the lips and to track them throughout the speech sequence.

In the following sections, we will present a new hybrid approach of lip feature extraction. Our approach applies in the first stage the active contour method to automatically localize the lip feature points in the speaker's face. In the second stage, we propose a spatial-temporal tracking method of these points based on the Freeman coding directions and on voting techniques. This POI tracking will carry out visual information describing the lip movements among the locution video sequence. Finally, this visual information will be used to classify and recognize the uttered viseme.

## 3. ALiFE: Lip POI Localization and tracking

In this phase, we start with the localization of the external contours of the lips on the first image of the video sequence. Then, we identify on these contours a set of POI that will be followed throughout the video locution sequence.

Thus, there are two problems: (1) the lip and POI localization, and (2) POI tracking in video sequence. The details of our approach are presented in the following sections.

### 3.1 Lip and POI Localization

Our approach for lip POI localization is to proceed first by detecting a lip contour and secondly by using this contour to identify a set of POI. One of the most efficient solutions to detect lip contour in the lip region, is the active contour techniques, commonly named "Snakes" [12], and [13]. This technique appeared in the mid 80's under the conjoined works of Kass, Witkin and Terzopoulos in physical constraint model and picture treatment [9]. This method meets a lot of successes thanks to its capacity to mix the two classic stages of detection of contours (extraction and chaining).

The active contours (or snakes) are deformable curves evolving in order to minimize functional energy, which are associated to them [9]. They move within the image of an initial position toward a final configuration that depends on the influence of the various terms of energy. The snake's energy consists of an internal energy named regularization or smoothing energy and an external energy of data adequacy.

The snake detection or active contour based method consists of placing around a detected shape an initial line of contour. This line deforms itself progressively according to the action of several strengths that push it toward the shape.

The implied strengths are derived from the three following energies associated to the snake:

- An exclusive energy due to the contour of the shape, called internal energy: $E_{int}$.
- A potential energy imposed by the image: $E_{ext}$. It is this energy that attracts the snake line toward the real contours present in the image.
- An $E_{cont}$ energy that expresses some supplementary constraints that can be imposed by the user.

Our objective is to localize different POI of the lips of speaker's face in the first frame video sequence (Figure 2). For the definition of these different energy terms, it is necessary to take into consideration this objective in our description.

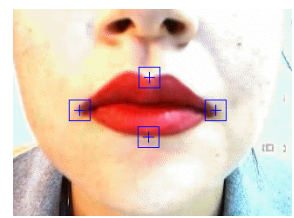

**Figure 2.** The Different POI for the extraction stapes

The core of our contribution is that active contour technique is applied just on the first frame of the video sequence and the addition of the $E_{cont}$ based on the direction of the snake evolution. In what follows, we present how we will adapt these terms of energies to our problem. The details of our lip localization approach are accessible in [31].

The three above energies can be defined as follows. We consider that our snake is composed of (*n*) *Vi* points with (i≤n), and that "*s*" is the parameter of spatial evolution in the contours picture, for example the curvilinear abscissa.

• The internal Energy:

$E_{int}$ is going to depend only on the shape of the snake. It is a regularity constraint curve. We calculate it according to Equation 1.

$$E_{int} = (a(s)*|V'(s)| + (b(s)*|V''(s)|)) \quad \text{(Equation 1)}$$

*Where a* and *b* are respectively the weights of the first and second derivative $V''$ and $V'$. We will adjust *a* and *b* to find a flexible contour (that will be able to wedge on the corners and the sharp angles: the corners and the Cupidon-bow) and a very regular contour that will follow the contour without clinging to a "*false alarm*".

• A potential energy imposed by the image:

$E_{ext}$ is characterized by a strong gradient depicted by Equation 2.

$$E_{ext} = -|\nabla I(x,y)|^2 \quad \text{(Equation 2)}$$

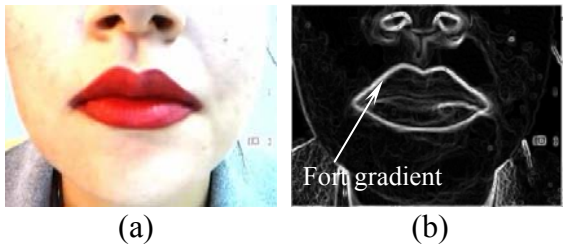

**(a)**          **(b)**

**Figure 3.** (a) Original image (b) gradient image

• The constraint energy:

$E_{cont}$ is often defined by the user, according to the specificities of the problem. One of the cores of our contribution is the definition of $E_{cont}$. For us, $E_{cont}$ aims at pushing the evolution of the snake toward the gravity centre G $(x_g, y_g)$ of the active contour (Figure 4). It represents the Euclidian distance between G and $V_i$ computed as follows:

$$E_{cont} = \sqrt{((x_s - x_g)^2 + (y_s - y_g)^2)} \quad \text{(Equation 3)}$$

With $(x_s, y_s)$ and $(x_g, y_g)$ the respective Cartesian coordinates of snake's points (s) and gravity center of the snake (G).

The principal goal of this energy is therefore ensure the evolution of the snake in the picture zones having a weak gradient.

Finally, the total energy of the snake: $E_{tot}$

The total energy in one point $V_i$ of snake ($Ei_{tot}$), is calculated as follows:

$$Ei_{tot}(V_{i-1}, V_i, V_{i+1}) = E_{int}(V_{i-1}, V_i, V_{i+1}) + E_{ext}(V_i) + E_{cont}(V_i)$$

*(Equation 4)*

Therefore the total energy of the snake E $_{tot}$ can be computed by the following equation:

$$E_{tot}(V_{i-1}, V_i, V_{i+1}) = \sum_{i=1 \to n}(Ei_{tot})$$

$$= \sum_{i=1 \to n}(Ei_{int}(V_{i-1}, V_i, V_{i+1}) + Ei_{ext}(V_i) + Ei_{cont}(V_i)).$$

*(Equation 5)*

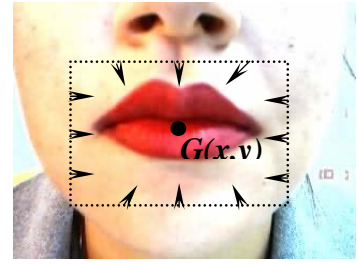

**Figure 4**. Snake directional energy principle

After the definition of the terms of the active contour energy, the snake is going to evolve progressively in order to minimize its total energy $E_{tot}$. We will stop the snake progression, when $E_{tot}$ reaches its minimal value or until attending a fixed number of iterations. For our experimentation, we have used the first method (Figure 5).

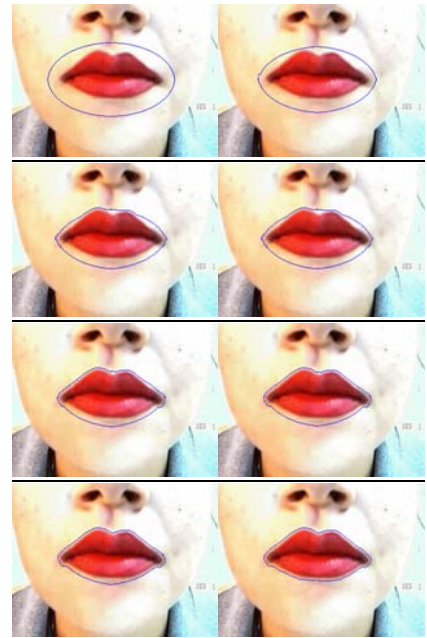

**Figure 5.** Snake evolution according to the energy minimisation principle

(The snake progression will be stopped when the $E_{tot}$ reaches the minimum value)

Once the external contours of the lips are extracted, we proceed to the detection and the initialization of the different POI. (We employ the word initialization because these POI will be the entries of the tracking step).

Here we intend to employ a technique of projection (horizontal and vertical) of the various points of the snake, to detect different POI. More precisely, the maximum projection on the horizontal axis indicates the position of two corners of the lips and the maximum's projection on the vertical axis indicate the position of the lower lip and the three points which form the arc of Cupid. Figure 6 illustrates this localization process.

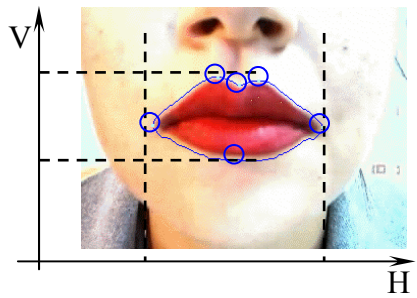

**Figure 6.** Points of interest detection by the projection of final contour on horizontal and vertical axis (H and V)

## 3.2 POI Tracking in a Video Sequence

The problem of POI tracking (in our context each POI is defined by a block of size w*w pixels) is to detect these POI in the successive images of the video sequence. This problem consists of looking for the block (*j*) on the image (*i*) which has the maximum of similarity with the block (*j*) detected on the image (*i-1*) knowing that (*i*) is the number of image in the video sequence and (*j*) is the number of block which define the different POI. Figure 7 shows an example of POI tracking on different images of the same sequence.

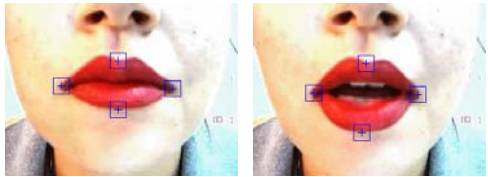

**Figure 7.** Example of tracking POI on Successive image: (a) First initialized image, (b) Detection of the different POI on an image of the same sequence

In the literature we find several methods for searching or tracking "Pattern" in a video sequence [15, 16, and 19]. Each of these approaches has been advocated by various authors, but there are fewer comparisons among them [17 and 18]. Generally, these different methods are confronted by the dilemma of high and low precision (it is a very delicate problem, and it is often very difficult to find the compromise between these two criteria). Finally, we can distinguish several disadvantages by the use of these methods for the Template Matching. We resume these disadvantages in the following three points:

− If the energy of the image varies with the position, the search for the model can fail. For example, if the correlation between the regions looked for and the model is well inferior to the correlation between the model and the effect of luminance.
− These methods depend enormously on the size of the model.
− They are sensitive to the abrupt luminance variation that can occur in video sequence. This is a serious disadvantage for our problematic because the lip movement can cause a big variation of luminance in the mouth region.

Therefore, we clearly notice that this search method of model is applicable in several cases, but we judge that it is not sufficient and convincing in our problematic, where the details are very important, the movements are rapid and the luminance variations are very susceptible (especially for the POI that concerns the inferior lip). This lead us to decided on taking a different approach of POI tracking in our study [29].

Our approach of POI tracking is based on the Template Matching technique, moreover, it take the spatial-temporal indices of the video into account. The principle of this approach consists of seeking in a gray level image gi(x, y) the most similar block to the block pattern forming a point of interest (POI) defined in section 3.2.

The originality of our technique of labial-movement tracking lies in the case of being limited to a set of POI. Our algorithm of tracking is based on two principle steps:

− The POI tracking is done in the different directions of the Freeman coding to localize the candidate points describing the potential POI movements;
− In the second steps, a vote technique is used to identify among all the candidate points, the one that corresponds better to the origin POI.

### 3.2.1. POI Tracking based on The Freeman coding:

In order to follow better the lips movements, we suppose in the present work that the head of the speaker is still with respect to the camera. The head immobility problem can be resolved by using a fixed camera. Indeed, today there are video-conference cameras to capture a centered shot on the labial zone. Figure 8 shows the headset of the Labiophone project on a synthesis clone conceived by Ganymedia [9].

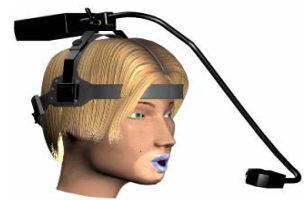

**Figure 8.** Headset of the Labiophone project on a synthesis clone conceived by Ganymédia [9]

With this assumption, we avoid the camera movement problems and we focus the objective of our tracking algorithm on the lip movements which are complex, fast and especially in all the directions. These different directions can be easily represented by the directions of the Freeman coding whose principle consists of presenting the connection of the successive pixels according to figure 9.

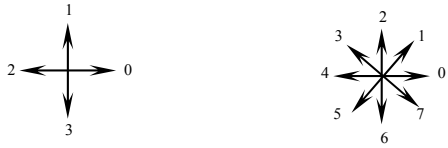

**Figure 9.** Different directions of Freeman coding

So if $P_n$ is a POI and $P_m$ is a point describing the potential movement of $P_n$, then the displacement of $P_n$ to $P_m$ can only be done in the 8 possible directions of the coding of Freeman.

Nevertheless, in order to increase the precision of our technique of POI movement predictions, we spread our research of the candidate points of POI to the set of the $P_m$ points being located in one of the directions of the Freeman coding on a ray of R points (R the number of block for every direction). In our experimentation R is set up to the value 10. Figure 10 illustrates this principle.

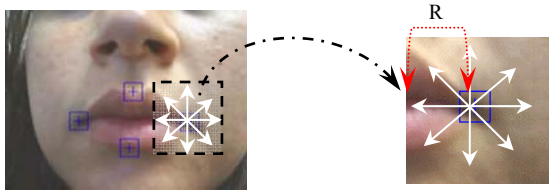

R: number of blocks for every direction

**Figure 10.** Different research directions according to the coding of Freeman on a ray of R points

### 3.2.2. Technical research and voting principle:

Another contribution of our tracking approach is through its use of a voting system for the selection of the most similar block to the pattern block. Two major stages constitute our voting system:

− The first step consists in calculating the list of the measures (D $\{b_n (x, y)\}$) in gray level one for (n) candidate blocks with the model to search for in every potential point. All the D $\{b_n (x, y)\}$ values will be saved in an accumulator that will be treated in a second part of our approach (Figure 11).

| Accumulator | | |
|---|---|---|
| D$\{b_1(x_1, y_1)\}$ | … | D$\{b_1 (x_{w^*w-1}, y_{w^*w-1})\}$ | D$\{b_1 (x_{w^*w}, y_{w^*w})\}$ |
| ⋮ | ⋮ | ⋮ |
| D$\{b_n(x_1, y_1)\}$ | … | D$\{b_n(x_{w^*w-1}, y_{w^*w-1})\}$ | D$\{b_n (x_{w^*w}, y_{w^*w})\}$ |

w: Block Size

**Figure 11.** 'Accumulator': Table of luminance measure for the different candidate Blocks

− In the second step, we add to our accumulator new columns where we will stock the number of voices for each block and we will vote at the end for the block having maximum voices (Figure 12).

− To calculate the luminance variation, we can use any measure of similarity (SSD, NCC …). For our case we calculate it as follows:

| Accumulator | | | Nb. voice |
|---|---|---|---|
| D$\{b_1(x_1, y_1)\}$ | … | D$\{b_1 (x_{w^*w}, y_{w^*w})\}$ | 13 |
| | | | |
| D$\{b_m(x_1, y_1)\}$ | … | D$\{b_m(x_{w^*w}, y_{w^*w})\}$ | **80** |
| | | Block elected for the research model | |
| D$\{b_n(x_1, y_1)\}$ | … | D$\{b_n(x_{w^*w}, y_{w^*w})\}$ | 26 |

**Figure 12.** 'Accumulator': Table of luminance measure with the vote principle

$$D\{b_n (x,y)\} = |g_i(x,y) - f_{i-1}(x,y)| \qquad \textit{(Equation 6)}$$

Where:

− D $\{b_n (x, y)\}$: Luminance variation Measure for the pixel (x, y) of block $b_n$ of the POI $_n$ with the respective pixel of the original POI.

− i: Number of image of the sequence.

− $g_i (x, y)$: Luminance of the model looked for.

− $f_{i-1} (x, y)$: Luminance of the image in which we make the research of the model.

− Votes are attributed according to the following algorithm :

```
For i = 1 To w*w
  Do
    Min ← Acc[i][1]
// Initialization of the minimal distance on a column
// Research of the minimal distance for the column i
// 8 the number of different directions of Freeman coding
// R the number of blocks for every direction
// w: Block Size
    For j = 2 To 8 * R
      Do
        If (Acc[i][j]<min)
          Min ← Acc[i] [j]
        EndIf
    EndDo

// Attribution of the voices for blocks having the
//minimal distance for the column I for every direction

    For j = 2 To 8 * R
      Do
        If (Acc[i][j]=min)
          Acc [w*w+1] [j] ← Acc [w*w+1] [j] +1
        EndIf
    EndDo

EndDo
```

With this method:
- we highlight all the details of the model that we are looking for,
- the result is independent of the size of the block,

- we diminish the noise effect; the latter will not influence the entirety of the candidate blocks.

In fact, thanks to our vote principle, the noise will affect only some voices, without affecting the whole result.

# 4. ALiFE: Lip Feature Extraction and classification : Application in French vowels

In this section, we present the different visual descriptors which we use for the characterization of the labial movements. These visual descriptors will be the entries and the only information on which the recognition phase will depend.

## 4.1. Lip Feature Extraction

The choice of the syllabic descriptors must be relevant and accurately describing the movement of each viseme in a corpus. In what follows, we will consider three physical natural descriptors: the horizontal distance between lip corners (DH), the vertical distance between the lower and the upper lip (DV) and the dark surface (Dark Area) inside the mouth (DA). The extraction of these descriptors will be based on the tracking of the four points already presented in section 4. Figure 13 illustrates an example of extraction of these distances on a speaker of our corpus.

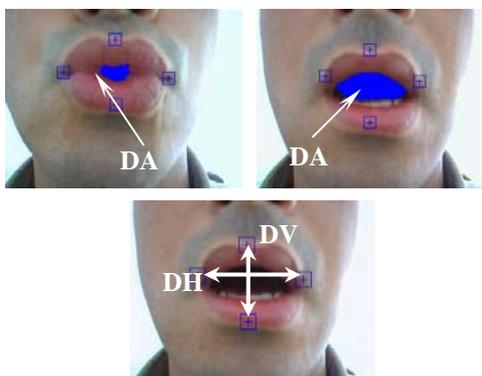

**Figure 13.** Different descriptors for extraction to the recognition stage

The variation of the vertical distance from the upper and lower lip gives a clear idea on the opening degree of the mouth during the syllabic sequence. This measure is very significant in the recognition of the syllables containing the vowels which open the mouth for example /ba/ (Figure 14). The variation of the horizontal distance between right and left lip corners describes the stretching intensity of the lips during the locution sequence. This measure is very significant for the recognition of the visemes containing vowels which stretch the mouth for example /bi/ (Figure 14). Finally, the DA descriptor, in spite of its irregular appearance, the dark surface is a relevant descriptor of the labial movement's characterization. In the following section, we justify the relevance and the intelligibility of this descriptor, as well as the method of extraction and the measure which we calculate for the recognition phase.

### 4.1.1. Intelligibility of Dark Area

Speech is the result of vocal combinations that have a symbolic value in a language. The speech happens after a stop in the expiry. This is a compression of the intra-thoracic air with closing of the glottis, then opening of the glottis and emission of the air, thus we hear the voice. The air expelled by the lungs crosses the larynx where it is put in vibration by successive opening/closing of the vocal cords or, more probably, by the undulation of the mucous membrane which covers them. More precisely, according to the configuration that the mouth organs can take, the air ejected by the lungs until outside, will pass through these organs which will produce the speech. This information gave us the idea to develop the measure, named dark area (DA). The dark area will defines the inter-labial surface from which the air will be ejected towards the outside of the mouth. We will see in our experiments that this descriptor is very relevant and constitutes discriminating criteria for the configuration of several visemes like /bou/ (Figure 14).

### 4.1.2. Extraction of Dark Area

To extract the dark pixels which are inside a mouth, we will seek these pixels in the region of interest (ROI) described by a polygonal form (figure 14). This region is formed by the four POI defined in section 4.

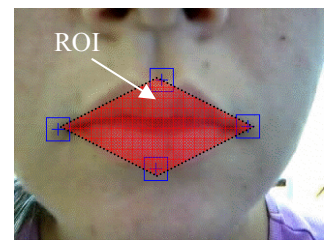

**Figure 14.** ROI: Region of Interest

The main problem is to separate between the dark and non-dark pixels. Then it is a question of finding a method which can operate at various conditions of elocution sequence acquisition, different configurations and colours of the vermilion (which is not regular for all speakers). For this, we propose an extraction of dark areas method with an adaptive threshold. The threshold ($S_{dark}$) will be calculated according to the following equation:

$$S_{dark} = \alpha \times \frac{\sum_{i=1}^{n} I(x_i, y_i)}{n} \quad \text{(Equation 7)}$$

*With n the number of pixel within the ROI.*

$\alpha$ is a coefficient which we fixed at 0.3 according to experimental results that we carried out on our audio visual corpus. The discrimination between the dark and non-dark pixels is done according to the following equation:

$$\begin{cases} \text{if } I(x,y) \le S_{dark} \quad \text{dark pixel} \\ \\ \text{else} \\ \\ \text{non} - \text{dark pixel} \end{cases} \quad \textit{(Equation 8)}$$

Figure 15 presents the results of the dark area extraction for various speakers with distinct colours and under different lighting conditions. Finally, we obtain, for each image of the elocution sequence, the number of dark pixels inside the ROI. Infact, the number of dark pixels is not so discriminating between the various configurations of each visemes. The *spatial position* of these pixels inside the ROI is more interesting and more relevant. To develop this criterion (spatial position), we proceed by a weighting of the dark pixels in their position compared to the ROI midpoint. The values of the dark area feature (V [DA]) will be calculated through the following rule:

$$V[DA] = \sqrt{(X - X_c)^2 + (Y - Y_c)^2} \quad \textit{(Equation 9)}$$

*With $X_c$ and $Y_c$ Cartesian coordinates of the ROI midpoint.*

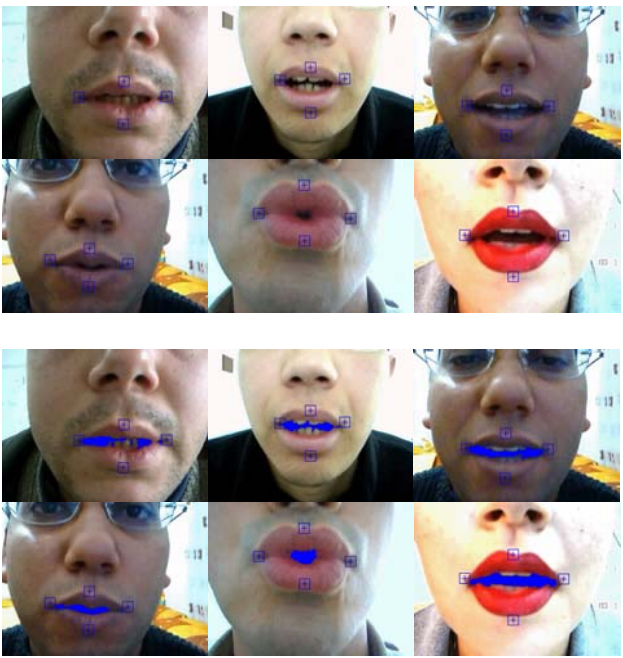

**Figure 15.** Result of the dark area extrzaction with various speakers under different lighting conditions (From left to right and from top to bottom the images without detection then the same images with the dark area detection)

With such an approach, we exploit the density of the spatial position of the dark pixels inside the ROI.

## 4.2. Lip feature classification and recognition

The robustness of a speech recognition system depends largely on the relevance of the descriptors and the training stage. In addition, a great number of data is necessary to ensure an effective training of the system. The experiments which we carry out on test data, different from the training data, make it possible to characterize the performances of the ALiFE system. That is why the development of recognition systems imposes the use of a considerable size of data. In the literature, the audio-visual or visual corpuses available for speech recognition are very rare [3]. Indeed, the constitution of such corpus posed material and storage problems. In our case, the viseme recognition, using an existing corpus becomes increasingly difficult because, on the one hand we focus on strictly visual data and on the other hand the nature of the speech unit to recognize which is the viseme. Thus, we were obliged to build our own audio-visual corpus.

In this section, we present our own audio-visual corpus and we detail our lip feature classification method. We will consider in the classification the three features defined in section 4.1.

### 4.2.1. Viseme Corpus presentation:

The ALiFE prototype is evaluated with multiple speakers under natural conditions. We have created a specific audio-visual (AV) corpus constituted by the different French vowels. Our AV corpus is composed of 42 native speakers, of various age and sex. The capture is done with one CCD camera; the resolution is 0.35 Mega of pixels and with 25 frames/s (fps). This cadence is widely enough to capture the major important lip movement.

In the French language, we note an identical lip movement for a group of vowels, but also a different lip movement for other groups. Table 1 illustrates the different lip configurations for all the French language vowels. We remark four different lip movements on this table, but we note a confusion zone between the groups "AN" and "O". Thus we can distinguish three differentiable lip movements for French vowels: group A (Opening movement), group O (Forward movement) and group I (Stretch movement) (Tab. 1).

The specificity of our corpus consists of syllable sequences that are visually differentiable uttered by female and male speakers: /ba/, /bi/ and /bou/. We justify the choice of the three treated cases in the following section.

The articulator gestures for vowels are generally stationary but their configurations required for a particular vowel are not very constrained (Tab. 1). The production of vowels in natural speech is generally more lax and depends on co-articulation and stress. Furthermore, the speaker can vary the visible configuration of many vowels without changing their auditory characteristics. Stevens' [22] classical works in the quantitative description of vowel articulation affirm that possible range of formant frequency combinations for producing vowels can then be obtained by varying as few as three parameters:

(a) Distance between the glottis and the maximum constriction

(b) Cross-sectional area at the constriction

(c) Ratio of area of lip opening to the length of lip passage

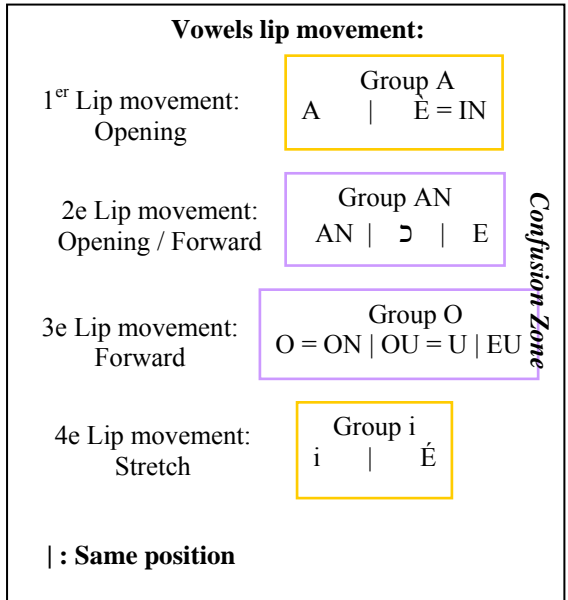

**Tab.1:** Lip movement for French Vowel

The main problem here is that out of these three parameters, only the third one (c) is clearly visible, which demonstrates the difficulty of visual vowel recognition. Since we are working on a syllable /CV/ recognition system, we judge that it is reasonable to evaluate in a first time our ALiFE beta version prototype on vowels and especially since we have conceived the (DA) feature (described in section 4) to compensate the no visibility of the two parameters (a) and (b) for producing vowels. Just characterizing the vowel production, we certainly think to enlarge our ALiFE recognition system to cover the totality of the French language visemes. But in this case, we predict that it is imperative to add other features to describe perfectly the visual production of consonants. However, this addition will not affect the structure of our ALiFE recognition system.

In the next section, we detail the different stages of our new approach for classification and recognition of these visemes.

### 4.2.2. ALiFE classification and recognition:

Our recognition system ALiFE, like the majority of recognition systems, is composed of two sub-systems, one for training and another one for recognition. The training stage consists of building the recognition models for each viseme in our corpus. In the second stage, we classify the viseme descriptors by the comparison with the models built during the first stage. The final objective is the recognition of the viseme. The details of our new approach for viseme classification and recognition are presented in [30].

### a) Training Sub-system:

The principle goal of this stage is to construct the different Syllable Descriptive Templates (SDT) that will be the basis of the recognition sub-system. For example, if we take the syllable /ba/, we must create three templates for the horizontal variation (DH), vertical opening distance (DV) and the dark region variation rate (DA) described at section 4.1.

After the POI tracking stage (section 4), we obtain three vectors describing the features variation belonging to the speech sequence, exp. V [DH] (Figure 17).

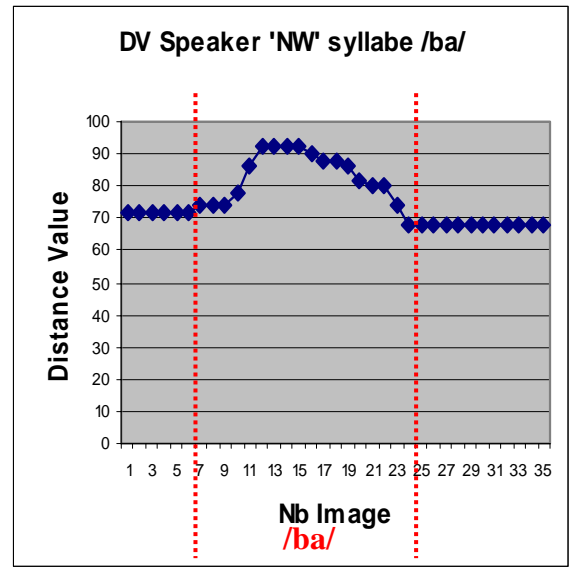

**Figure 16**. Tracking of Vertical variations Distance (*DV*) with syllable /ba/ for the speaker "NW"

We see on the curve of figure 17, that the DV value remains steady within the frame 1 until 7 and 25 until 35, these values correspond to a pause in the start and the end of the speech sequence. These DV values will be ignored, in order not to influence the final recognition template (Figure 16).

On the other hand, we note that the syllable sequence duration is not necessarily unvaried for all speakers. Moreover, we do not have the same mouth size for all people. So, to construct a robust recognition template it is important to apply a spatial-temporal normalization on the distance variation tracking curves (Figure 17).

During this operation, we have maintained the same appearance of the tracking variation curve by the use of linear interpolation operations between the curve points. The goal of this process is to assure the representation of each tracking variation curve by the use of the same number of points ($\omega$). In our case, $\omega$ is fixed to *10*.

The choice of the ($\omega$) value is not at random, but the number that we choose must be widely sufficient to give a faithful representation of a syllable tracking variation curve. Thus, the value of ($\omega$) will depend on the duration of the speech unit to recognize. For us, since we work on syllables, the image sample number ($\omega$) will depend on the syllable elocution duration.

In the French language the study of the temporal structuring of a corpus largely depends on the conditions of speech flow. Pasdeloup [20] describes the duration of the syllable in various rhythms. We notice that the average duration of syllable elocution is 400 ms (Figure 17).

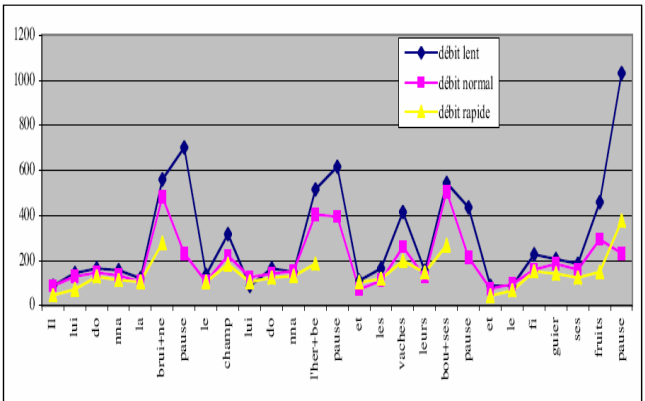

**Figure 17.** Syllabic duration in French language (for three flow conditions: slow, normal and fast)

Therefore, if we captured the elocution sequence with 25 frames/s (fps) and the syllable duration is 0.4 seconds then the number of frame that gives a faithful representation of labial movements is equal to 10. Generally, if we have speech unit duration ($D$) and a frame capture cadence ($Cd$) the value of the ($\omega$) will be calculated by the following equation:

$$\omega = Cd \times D \qquad (Equation\ 10)$$

For the spatial normalization common factor, we have opted for the initial horizontal and vertical distance for features ($DH$, $DV$) and the maximal the maximal value of the dark pixel rate in the speech sequence for the feature ($DA$). Figure 18 shows the result curve of the spatial-temporal normalization on the vertical distance variation curve with the syllable /ba/.

We have mentioned above that the principle aim of this training sub-system is to build three-feature recognition templates for every viseme in the corpus. These templates will be generated from the average of all retrieved variation curves for every syllable feature.

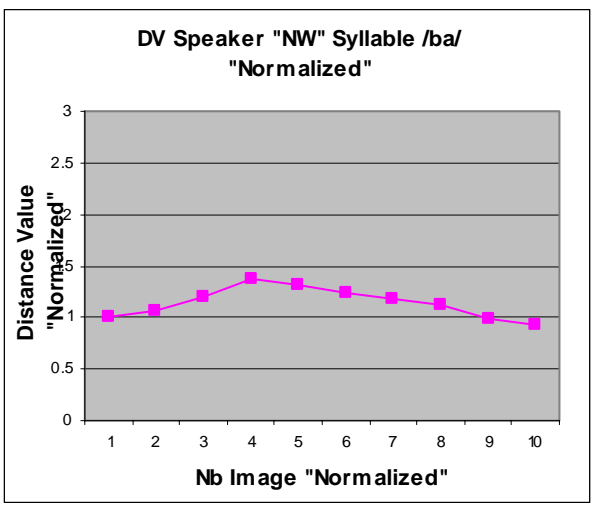

**Figure 18.** Normalization of the Vertical Distance variations ($DV$) with syllable /ba/ for the speaker "NW"

Figure 19, shows the generated syllable template ($SDT$) for the syllable /ba/ with the vertical variation distance feature. The $ID1$ to $ID30$ describes all vertical distance variation obtained form 30 native speakers.

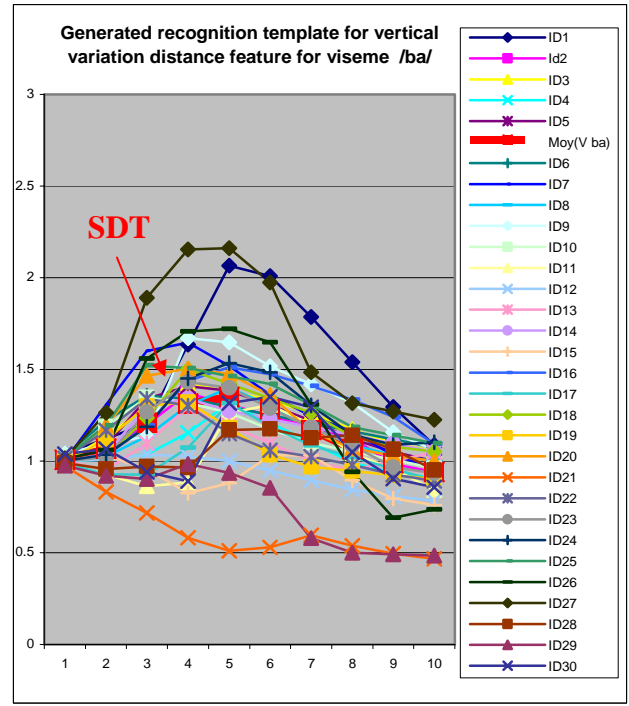

**Figure 19.** Generation of the Vertical Distance variation Template ($SDT$) with syllable /ba/.

### b) Recognition Sub-system:

After accomplishing the POI localization and tracking, the extraction of a feature and the normalization of the feature vectors, the input speech sequence is characterized by three feature vectors $fVi$ ( $i$ : feature index, $i \in [1..3]$). Each normalized $fVi$ is then characterized by ($\omega$) points (with $\omega=10$ as described above see section 4.2.2).

Consequently, our recognition system consists of comparing each $fVi$ with its respective $SDT$ as shown in figure 20.

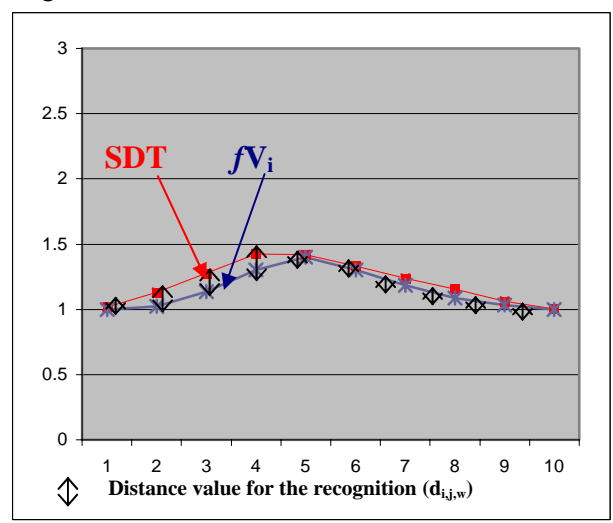

**Figure 20.** Distances to calculate between the $SDT$ and the $fV_i$ with $DV$ and syllable /ba/

This comparison generates (ω) distances $d_{ij1}, \ldots d_{ij10}$ ( i : feature index, i ∈ [1..3], j : syllable index, j ∈ [1..3] ). Each obtained distance is used as an input to our Neural network after the f transformation function given by (Equation 12):

$$e_{ijw} = f(d_{ijw}) = \frac{1}{\left(1 + \alpha * d_{ijw}\right)} \quad \text{(Equation 11)}$$

$\alpha$ : Constant coefficient for distance $(d_{i,j,w})$ used as a scaling parameter.

Hence, our recognition sub-system is composed of ($i*j$) neural network: precisely, we define for every feature ($i$) and syllable ($j$) a neural network composed of (ω) input layers (identity) and one output layers (Figure 21).

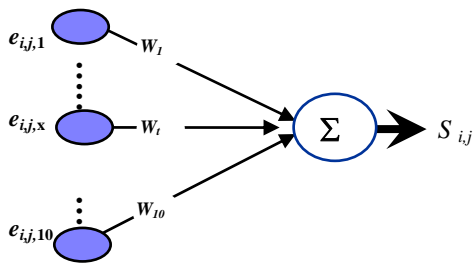

i: Feature index
j: Syllable index

**Figure 21.** Neural Network architecture for distance vector classification

All the Neural Network outputs ($S_{i,j}$), are used to compute the probability $P(SYL_j)$ of speech sequence which corresponding to the syllable ($SYL_j$). The recognition of the syllable is based on the posteriori maximum probability (PMP) calculated by equation 13.

$$P(SYL_j) = \arg\max_j P(SYL_j / O) \quad \text{(Equation 12)}$$

With O the number of observations related to the different descriptors ($F_i$).
We add a weight coefficient ($C_i$) for every descriptor with each syllable. This coefficient will indicate the influence rate for the chosen descriptor ($i$) with the recognition of a syllable ($j$). So we can detail the equation 13 in the following way:

$$P(SYL_j / O) = \sum_i C_i * P(SYL_j / F_i) \quad \text{(Equation 13)}$$

The probability $P(SYL_j / F_i)$ can be calculated according to the Bayes Theorem:

$$P(SYL_j / F_i) = \frac{P(F_i / SYL_j) * P(SYL_j)}{P(F_i)} \quad \text{(Equation 14)}$$

In equation 15, we can ignore the denominator $P(F_i)$ since it is the same for all the descriptors. The probability $P(SYL_j)$ is supposed to be constant for all the syllables of the corpus. Consequently, the conditional probability of equation 15 can be calculated in the following way:

$$P(F_i / SYL_j) = \frac{S_{i,j}}{\sum_{i=1 \to 3} S_{i,j}} \quad \text{(Equation 15)}$$

With $S_{i,j}$ neural network output.

Finally, we have to recognize the syllable which has the highest probability $P(SYL_j / O)$.

## 5. Experimental Results:

This section is composed of two experimental parts: a first experimentation to compare our spatial-temporal lip tracking method with other techniques, and a second experimentation for the evaluation of our ALiFE system for the visual recognition of each viseme present in our corpus. The rates of recognition of each viseme as well as a matrix of confusion between these visemes will be shown.

### 5.1. ALiFE lip tracking experimentation

SIMI°MatchiX is an image processing software for automatic movement tracking. The pattern matching algorithm can be utilized with video clips to automatically track user-defined patterns [21]. The approach used by the SIMI°MatchiX is the Normal Cross Correlation (NCC). We use this software to situate our 'Spatial-temporal voting algorithm according to other model search techniques. The experimental results are shown in Figure 22.

In these results, we notice deficiencies of the 'SIMI°MatchiX' system especially for the movements of the lower lip due to the luminance variations which are very important in this zone.

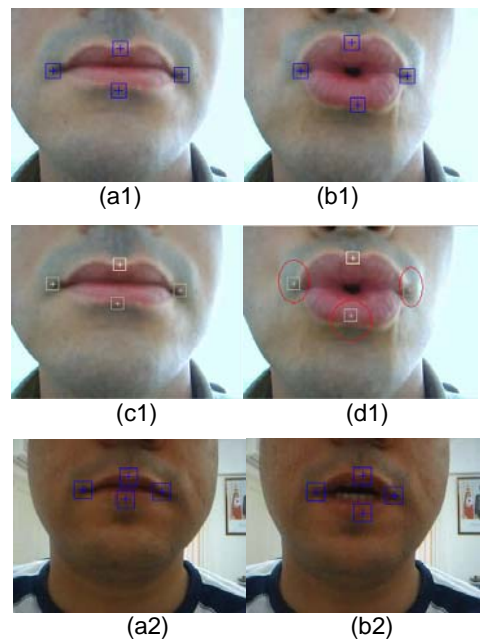

(a1)            (b1)

(c1)            (d1)

(a2)            (b2)

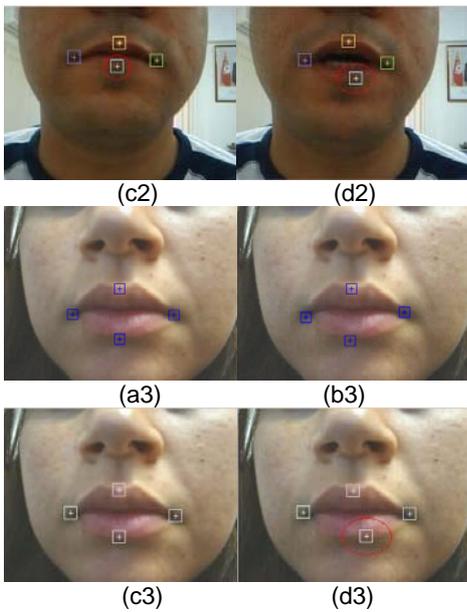

(c2)    (d2)

(a3)    (b3)

(c3)    (d3)

**Figure 22.** Experimental Results for different speakers: (a) (b) followed by vote algorithm and (c) (d) followed by software 'SIMI°MatchiX'

## 5.2. ALiFE viseme recognition experimentation

We have evaluated the ALiFE System with our audio-visual corpus. We have used 70% of our audio-visual corpus for training stage and 30% for recognition. The experimental results are presented in Tab. 2, Tab. 3 and Tab.4.

|  | ba | bi | bou | Recognition Rate |
|---|---|---|---|---|
| **ba** | 19 | 7 | 4 | **63.33%** |
| **bi** | 4 | 22 | 4 | **73.33%** |
| **bou** | 3 | 2 | 25 | **83.33%** |

**Tab.2:** Recognition rate of French Vowel Training Stage

|  | ba | bi | bou | Recognition Rate |
|---|---|---|---|---|
| **ba** | 7 | 4 | 0 | **63.64%** |
| **bi** | 3 | 8 | 0 | **72.73%** |
| **bou** | 0 | 2 | 9 | **81.82%** |

**Tab.3:** Recognition rate of French Vowel Recognition stage

|  | Recognition rate | |
|---|---|---|
|  | Training Stage | Recognition Stage |
| **ba** | 63.33% | 63.64% |
| **bi** | 73.33% | 72.73% |
| **bou** | 83.33% | 81.82% |
| **Recognition rate** | **73.33%** | **72.73%** |

**Tab.4:** Recognition rate of French Vowel (training and recognition stage)

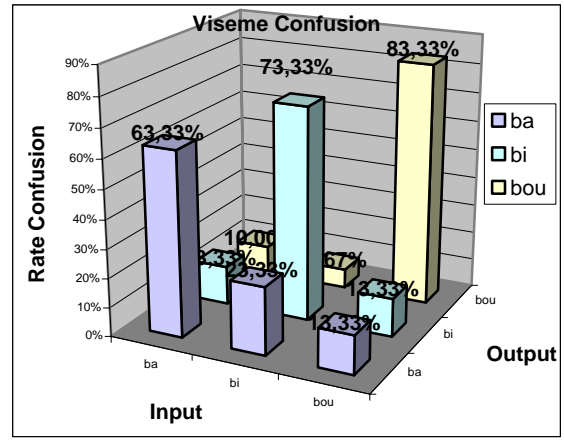

**Figure 23.** Experimental results of viseme confusion Training Stage

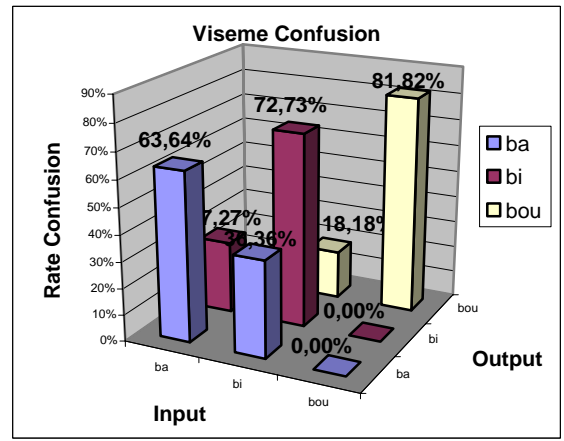

**Figure 24.** Experimental results of viseme confusion Recognition Stage

In these results, we notice that there is a good recognition rate for the viseme /bou/ (thanks to the addition of the dark area feature), but it is not the case for the other visemes. This poor rate recognition is due to the big confusion rate (23.33%) between the visemes /ba/ and /bi/, as shown in Figure 23 and 24.

## 6. Conclusion and future work

Several experiments show that visual channel carries out useful information for speech recognition. Many works in the literature, from the oldest [1] until the most recent ones [3], have proved the efficiency of the visual speech-recognition systems, particularly in noisy audio conditions.

Our research use visual information for the automatic speech recognition. The major difficulty of in a lip-reading system is the extraction of the visual speech descriptors. In fact, to ensure this task, it is necessary to carry out an automatic tracking of the labial gestures. Lip tracking constitutes in itself an important difficulty. This complexity consists of the capacity to treat the immense variability of the lip movement for the same speaker and the various lip configurations between different speakers.

In this paper, we have presented our ALiFE beta version system of visual speech recognition. ALiFE is a system for the extraction of visual speech features and their modeling for visual speech recognition. The system includes three principle parts: lip localization and tracking, lip feature extraction, and the classification and recognition of the viseme. This system has been tested with success on our audio-visual corpus, for the tracking of characteristic points on lip contours and for the recognition of the viseme.

However, more work should be carried out to improve the efficacy of our lip-reading system. As a perspective of this work, we propose to add other consistent features (for example the appearance rate of tooth and tongue). We also propose to enhance the recognition stage by the adequate definition of the feature coefficients for each viseme (use of the principal component analysis ACP). Finally, we plan to enlarge the content of our audio-visual corpus to cover the totality of the French language visemes and why not to discover other languages.

# References

[1] Petajan, E. D., Bischoff, B., Bodoff, D., and Brooke, N. M., "An improved automatic lipreading system to enhance speech recognition," *CHI 88*, pp. 19-25, 1985.

[2] P. Daubias, "Modèles a posteriori de la forme et de l'apparence des lèvres pour la reconnaissance automatique de la parole audiovisuelle". *Thèse à l'Université de Maine France* 05-12-2002.

[3] R. Goecke, "A Stereo Vision Lip Tracking Algorithm and Subsequent Statistical Analyses of the Audio-Video Correlation in Australian English". *Thesis Research School of Information Sciences and Engineering. The Australian National University Canberra, Australia,* January 2004.

[4] McGurck et J. Mcdonald. "Hearing lips and seeing voice". *Nature*, 264 : 746-748, Decb 1976.

[5] I. Matthews, J. Andrew Bangham, and Stephen J. Cox. "Audiovisual speech recognition using multiscale nonlinear image decomposition". *Proc . 4th ICSLP, volume1, page 38-41,* Philadelphia, PA, USA, Octob 1996.

[6] U. Meier, R. Stiefelhagen, J. Yang et A. Waibe. "Towards unrestricted lip reading". *Proc 2nd International conference on multimodal Interfaces (ICMI)*, Hong-kong, Jan 1999.

[7] K. Prasad, D. Stork, and G. Wolff, "Preprocessing *video* images for neural learning of lipreading," *Technical Report CRC-TR-9326, Ricoh California Research Center,* September 1993.

[8] R. Rao, and R. Mersereau, "On merging hidden Markov models with deformable templates," *ICIP 95, Washington D.C., 1995.*

[9] P. Delmas, Extraction des contours des lèvres d'un visage parlant par contours actif (Application à la communication multimodale) ". *Thèse à l'Institut National de polytechnique de Grenoble,* 12-04-2000.

[10] G. Potamianos, H. P. Graft et E. Gosatto. An *Image* transform approach For HM based automatic lipreading". *Proc, ICIP, Volume III, pages 173-177, Chicago, IL, USA* Octb 1998.

[11] I. Matthews, J. Andrew Bangham, and S. J. Cox. A *comparaison* of active shape models and scale decomposition based features for visual speech recognition". *LNCS, 1407 514-528,* 1998.

[12] N.Eveno, "Segmentation des lèvres par un modèle déformable analytique", *Thèse de doctorat de l'INPG, Grenoble,* Novembre 2003.

[13] N. Eveno, A. Caplier, and P-Y Coulon, "Accurate *and* Quasi-Automatic Lip Tracking" , *IEEE Transaction on circuits and video technology,* Mai 2004.

[14] L. R. Rabiner. "A tutorial on Hidden Markov models and *selected* applications in speech recognition". *Proceedings of the IE*EE, 77(2) : 257-286, February 1989.

[15] P. Anandan, "A *Computational* Framework and an Algorithm for the Measurement of Visual Motion", *Int. J. Computer Vision*, 2(3), p. 283-310, 1989.

[16] D. I. Barnea, H. F. Silverman, "A class of *algorithms* for fast digital image registration", *IEEE Trans. Computers*, 21, pp. 179-186, 1972.

[17] J. Shi and C. *Tomasi*, "Good Features to Track", Proc. *IEEE Conf. on Computer Vision and Pattern Recognition*, 1994.

[18] B. D. Lucas *and* T. Kanade, "An Iterative Image Registration Technique with an Application to Stereo Vision", *IJCAI* 1981.

[19] A. R. Lindsey, "The Non-Existence of a Wavelet Function Admitting a Wavelet

Transform *Convolution* Theorem of the Fourier Type", *Rome Laboratory Technical Report* C3BB, 1995.

[20] V. Pasdeloup, "Le rythme n'est pas élastique : étude *préliminaire* de l'influence du débit de parole sur la structuration temporelle", *Département de Lettres, Université de Rennes 26 avenue Gaston Berger* CS 24 307, F-35043 Rennes, France.

[21] SIMI *Reality* Motion Systems, GmbH, D-85705 Unterschleissheim, Germany. www.simi.com/matchix

[22] K. N. *Stevens* and A. S. House. "Developpement of a quantitative description of vowel articulation". *Journal of the Acoustical Society of America,* 27: 484-493, 1955.

[23] Y. Nakata and M. Ando. "Lipreading Method Using Color Extraction Method and Eigenspace Technique", *Systems and Computers in Japan, Vol. 35,* No. 3, 2004.

[24] J. WU, S. Tamura, H. Mitsumoto, H. Kawai, K. Kurosu, and K. Okazaki. "Neural network vowel recognition jointly using voice features and mouth shape image". *Pattern Recognition,* 24(10): 921-927, 1991.

[25] E. D. Petajan, N. M. Brooke, B. J. Bischoff, and D. A. Bodoff. "An improved automatic lipreading system to enhance speech recognition". In E. Soloway, D. Frye, and S. B. Sheppard, editors, Proc. *Human Factors in Computing Systems, pages 19-25 ACM*, 1988.

[26] A. J. Goldschen, O. N. Garcia, and E. Petajan. "Continuous optical automatic speech recognition by lipreading." *In 28th Annual Asilmomar Conference on Signals, Systems, and Computer*, 1994.

[27] C. Bregler and Y. Konig. Eigenlips for robust "speech recognition. In Proc. IEEE Int. Conf. on Acoust.", *Speech, and Signal Processing, pages 669-672, Adelaide*, 1994.

[28] Y. Nankaku, K. Tokuda. "Normalized Training for HMM-Based Visual Speech Recognition". *Electronics and Communications in Japan, Part 3, Vol. 89, No. 11*, 2006.

[29] S. Werda, W. Mahdi and A. Benhamadou, "A Spatial-Temporal technique of Viseme Extraction: Application in Speech Recognition", *SITIS 05, IEEE, The International Conference on Signal-Image Technology & Internet–Based System, Yaoundé*, 2005.

[30] S. Werda, W. Mahdi, M. Tmar and A. Benhamadou, "ALiFE: Automatic Lip Feature Extraction: A New Approach for Speech Recognition Application ", *the 2nd IEEE International Conference on Information & Communication Technologies: from Theory to Applications - ICTTA'06 - Damascus, Syria.* 2006.

[31] S. Werda, W. Mahdi, and A. Benhamadou, "ALiFE: A New Approach for Automatic Lip Feature Extraction: Application in Speech Recognition", *in 3rd IEEE International Symposium on Image/Video communications over fixed and mobile networks, Yasmine-Hammamet, Tunisia*, 2006.

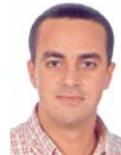
**Salah WERDA** is preparing a Doctorate degree in Computer Science at the *Faculté des Sciences Economique et de Gestion de Sfax*, Tunisia. he is a Teaching Assistant at the Higher Institute of Computer Science and Multimedia, Sfax, Tunisia. His research is about image and video processing and is focused on the visual speech recognition (Lipreading System).

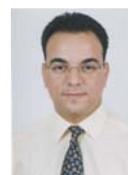
**Walid MAHDI** received a Ph.d. in Computer and Information Science from *Ecole Centrale de Lyon, France* in 2001. He is currently Assistant Professor at Higher Institute of Computer Science and Multimedia, at the University of Sfax, Tunisia. His research is about image and video processing.

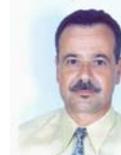
**Abdelmajid Ben-Hamadou** is Professor of Computer Science and the director of the Higher Institute of Computer Science and Multimedia, Sfax, Tunisia. He has been the director of the LARIS reserach laboratory since 1983. His research interests include automatic processing of Natural Language, object-oriented design and component software specification, and image and video processing.